\documentclass{article} 
\usepackage{iclr2015,times}
\usepackage{hyperref}
\usepackage{url}
\usepackage{amsmath,amssymb,bbold}
\usepackage{graphicx}
\usepackage{latexsym}
\usepackage{caption} 
\title{Modeling Compositionality with\\ Multiplicative Recurrent Neural Networks}

\author{
Ozan \.Irsoy \& Claire Cardie \\
Department of Computer Science \\
Cornell University \\
Ithaca, NY \\
\texttt{\{oirsoy,cardie\}@cs.cornell.edu} \\
}

\iclrfinalcopy 

\iclrconference 

\begin{document}

\maketitle

\begin{abstract}
  We present the multiplicative recurrent neural network as a general model
for compositional meaning in language, and evaluate it
on the task of fine-grained sentiment analysis. We establish a
connection to the previously investigated
matrix-space models for compositionality, and show
they are special cases of the multiplicative recurrent net. 
Our experiments show that
these models perform comparably or better than 
Elman-type additive recurrent neural networks and outperform
matrix-space models on a standard fine-grained sentiment analysis corpus. Furthermore, they yield comparable
results to structural deep models on the recently published Stanford Sentiment
Treebank 
without the need for generating parse trees.
\end{abstract}

\section{Introduction}
Recent advancements in neural networks and deep learning have provided fruitful applications
for natural language processing (NLP) tasks. One important such advancement 
was the invention of \emph{word embeddings} that represent a single word as a 
dense, low-dimensional vector in a meaning space~\citep{bengioNeural} from which numerous problems in NLP have
benefited~\citep{collobert-weston,collobertScratch}. The natural next question, then, was
how to properly map larger phrases into such dense
representations for NLP tasks
that require properly capturing 
their meaning.
Most existing methods take a compositional
approach by defining a function that composes multiple word vector representations into a phrase 
representation (e.g.\ \citet{mikolovnips}, \citet{socherRNTN}, \citet{ainur}).

Compositional \textbf{matrix-space models} \citep{rudolph2010,ainur}, for example, represent phrase-level
meanings in a vector space and
represent words as matrices that act on this vector space. Therefore, a matrix assigned to a word should
capture how it transforms the meaning space (e.g.\ negation or intensification). 
Meaning representations for longer phrases are simply computed as a multiplication of word matrices in sequential order (left-to-right, for English).
Their representational power, however, is accompanied by a large number of parameters --- a matrix for every word in the vocabulary. Thus, learning can be difficult.

But sequential composition of words into phrases is not the only mechanism for tackling semantic composition. 
\textbf{Recursive neural networks} \citep{pollack:raam}, for example, 
employ a structural approach to compositionality:  the composition function for a phrase operates on its two children
in a binary parse tree of the sentence. Single words are represented in a vector-space.
Different ways of defining the composition function 
lead to different variants of the recursive neural network.
In \citet{socher2011parsing}, 
a simple additive affine
function with an additional nonlinearity is used. The matrix-vector recursive neural 
network of \citet{socherMVRNN} extends this by assigning
an additional matrix to each word, similar to the aforementioned matrix-space models; and 
the composition function involves a matrix-vector
multiplication of sibling representations. More recently, \citet{socherRNTN} defines a bilinear tensor multiplication as the composition function --- 
to capture multiplicative interactions between siblings.

On the other hand, {\bf recurrent neural networks} (RNNs), a neural network architecture 
with sequential prediction capabilities, {\em implicitly} model compositionality 
when applied to natural language sentences.  Representation of a phrase
can be conceptualized as a nonlinear function that acts on the network's
hidden layer (memory), which results from
repeated function composition over the hidden layer and the next word in the phrase/sentence 
(see Section~\ref{sec:recurrent}). Unfortunately, it is possible that conventional 
additive recurrent networks are not powerful enough to accommodate some of
the more complex effects in language, as suggested in previous work on 
(multiplicative and additive variants of)  
recursive neural networks (e.g.\ \citet{socherRNTN}). 
More specifically, even though additive models can theoretically 
model arbitrary functions when combined with a nonlinearity,
they might require a very large number of hidden units, and learnability of large parameter
sets from data might pose an issue.

To this end we investigate the \textbf{multiplicative recurrent neural network} as a 
model for compositional semantic effects in language. 
Previously, this type of
multiplicative sequential approach has been applied to a character-level text
generation task \citep{RNTN}. In this work, we investigate its capacity for 
recognizing the sentiment of a sentence or a phrase represented as a sequence 
of dense word vectors. 
Like the matrix-space models, 
multiplicative RNNs are sequential models of language; and as a type of recurrent NN,
they implicitly model compositionality.  Like the very successful multiplicative recursive
neural networks, multiplicative RNNs can capture the same types of sibling interactions, but
are much simpler.  In particular, no parse
trees are required, so sequential computations replace the associated recursive computations
and performance does not depend on the accuracy of the parser.

We also show a connection between the multiplicative RNN and compositional matrix-space
models, which have also been applied to sentiment
analysis~\citep{rudolph2010,ainur}. 
In particular, matrix-space models are effectively a special case of multiplicative
RNNs in which a word is represented as a large ``one-hot'' vector instead of a dense small one.
Thus, these networks carry over the idea of matrix-space models from a one-hot sparse 
representation to dense word vectors. 
They can directly employ word vector representations, which makes them better suited for
semi-supervised learning given the plethora of word vector training schemes.
Multiplicative recurrent networks can be considered to unify 
these two views of 
distributed language processing --- the operator semantics view of matrix-space models in which a word is
interpreted as an operator acting on the meaning representation, and the sequential memory processing 
view of recurrent neural networks.

Our experiments show that multiplicative RNNs provide comparable or better performance than conventional
additive recurrent nets and matrix-space models in terms of fine-grained sentiment detection accuracy. Furthermore, although the absence of parse tree information puts an additional learning burden on
multiplicative RNNs, we find that they can reach comparable performance to the recursive neural network variants that require parse tree annotations for each sentence.

\section{Related Work}

\textbf{Vector Space Models.}
In natural language
processing, a common way of representing
a single token as a vector is to use a ``one-hot'' vector per token, with a dimensionality of the vocabulary
size. This results in a very
high dimensional, sparse representation. Additionally, every word is put at an equal distance to one another,
disregarding their syntactic or semantic similarities. Alternatively, a distributed representation maps a token
to a real-valued dense vector of smaller size (usually on the order of 100 dimensions). Generally,
these representations are learned in an unsupervised manner from a large corpus, e.g.\ Wikipedia. Various
architectures have been explored to learn these embeddings ~\citep{bengioNeural,collobert-weston,hlbl,mikolov}
which might have different generalization capabilities depending on the task ~\citep{turian}. The
geometry of the induced word vector space might have interesting semantic properties 
(\emph{king} - \emph{man} + \emph{woman} $\approx$ \emph{queen}) \citep{mikolov,mikolovnips}.
In this work, we employ such word vector representations as the initial input representation when
training neural networks.

\textbf{Matrix Space Models.}
An alternative approach is to embed words into a matrix space, by assigning
matrices to words. 
Intuitively, a matrix embedding of a word is desired in order to
capture operator semantics: the embedding should model how a word transforms meaning
when it is applied to a context. \citet{baroni2010} partially apply this idea to
model adjectives as matrices that act on 
noun vectors. In their theoretical work,
\citet{rudolph2010} define a proper matrix space model by assigning every word
to a matrix; representations for longer phrases are computed by matrix 
multiplication. They show that matrix space models generalize vector space models and
argue that they are neurologically and
psychologically plausible. \citet{ainur} apply this model to fine-grained sentiment
detection. \citet{socherMVRNN} use a structural approach in which every word
is assigned a matrix-vector pair, where the vector captures
the meaning of the word in isolation and the matrix captures how it transforms
meaning when applied to a vector.

\textbf{Compositionality in Vector and Matrix Spaces.}
Commutative vector operations such as addition (e.g.\ bag-of-words) or 
element-wise multiplication along with negation~\citep{widdows:2003:ACL} 
provide
simple composition schemes~\citep{Mitchell:Lapata:2010,zanzotto-EtAl:2010:PAPERS}. Even though they ignore the order of the words,
they might prove effective depending on the length
of the phrases, and on the 
task~\citep{mikolovnips}. 
Other models for compositional distributional semantics emulate formal semantics by representing functions as tensors and 
arguments as vectors (e.g.\ \citep{clark,coecke,grefenstette-et-al:2013:IWCS2013}) for which \citep{grefenstette-et-al:2013:IWCS2013} generalise the  tensor-learning approach of \citep{baroni2010}. 
More complex non-commutative composition functions can be modeled via sequential or
structural models of the sentence. In particular,
compositionality in recurrent neural networks can 
be considered as tranformations
on the memory (hidden layer) applied by successive word vectors in order. Recursive neural
networks employ a structural setting where compositions of smaller phrases into larger ones
are determined by their parent-children
relationship in 
the associated binary parse tree~\citep{socher2011parsing,socherMVRNN,socherRNTN}. In matrix space models, compositionality is
naturally modeled via function composition in sequence~\citep{rudolph2010,ainur}.

\textbf{Sentiment Analysis.}
Sentiment analysis has been a very active area among NLP researchers, at various granularities
such as the
word-, \mbox{phrase-}, sentence- or document-level~\citep{pang2008}. Besides preexisting work 
that tried to formulate the problem as binary classification, 
recently fine-grained approaches were explored~\citep{ainur,socherRNTN}.
Ultimately, the vast majority 
of approaches do not tackle the task compositionally, and in addition to bag-of-words features,
they incorporate engineered features to account for negators, intensifiers and contextual valence 
shifters~\citep{polanyi,wilson2005,kennedy,shaikh}.

\section{Preliminaries}

\subsection{Matrix-space models}

A matrix-space model models a single word as a square matrix that transforms a
meaning (state) vector to another vector in the same meaning space. Intuitively, 
a word is viewed as a function, or an operator (in this particular case, linear)
that acts on the meaning representation. Therefore, a phrase (or any sequence of words)
is represented as successive application of the individual operators inside the phrase.

Let $s = w_1, w_2, \mathellipsis, w_T$ be a sequence of words of length $T$ and 
let $M_w \in \mathbb{R}^{m\times m}$
denote the matrix representation of a word $w\in \mathcal{V}$ where $\mathcal{V}$
is the vocabulary. Then, the representation of $s$ is simply
\begin{align}
M(s) = M_{w_1} M_{w_2} \mathellipsis M_{w_T}
\end{align}
which yields another linear transformation in the same space. Observe that this 
representation respects word order (unlike, e.g.\ a bag of words). Note that even though
$M(s)$ is modeled as a linear operator on the meaning space, 
$M(s)$ as a function of $\{M_{w_i}\}_{i=1..T}$
is not linear, since it constitutes multiplications of those terms.

Applying this representation to a task is simply applying the function to an initial
\emph{empty} meaning vector $h_0$, which results in a transformed, \emph{final} meaning
vector $h$ that then is used to make a decision on the phrase $s$. In the case of 
sentiment detection, a sentiment score $y(s)$ can be assigned to $s$ as follows:
\begin{align}
y(s) = h^\top u = h_0^\top M(s) u \label{eq:matspa}
\end{align}

In such a supervised task, matrix-space model parameters $\{M_w\}_{w\in\mathcal{V}}, h_0, u$
are learned from data. $h_0$ and $u$ can be fixed (without reducing the representative power of
the model) to reduce the degrees of freedom during training.

\subsection{Recurrent neural networks}
\label{sec:recurrent}

A recurrent neural network (RNN) is a class of neural network that has recurrent connections, which allow a form of memory. 
This makes them applicable for sequential prediction tasks of arbitrary spatio-temporal dimension. They model the conditional
distribution of a set (or a sequence) of output variables, 
given an input sequence.
In this work,
we focus our attention on only Elman-type networks ~\citep{elman}.

In the Elman-type network, the hidden layer $h_t$ at time step $t$
is computed from a nonlinear transformation of the
current input layer
$x_t$ and the previous hidden layer $h_{t-1}$. Then, the
final output $y_t$ is computed using the hidden layer $h_t$.
One can interpret $h_t$ as an intermediate representation
summarizing the past so far.

More formally, given a sequence of vectors $\{x_t\}_{t=1..T}$, an Elman-type RNN
operates by computing the following memory and output sequences:
\begin{align}
h_t &= f(W x_t + V h_{t-1} + b)\\
y_t &= g(U h_t + c)
\end{align}
where $f$ is a nonlinearity, such as the element-wise sigmoid function,
$g$ is the output nonlinearity, such as the softmax function,
$W$ and $V$ are weight matrices between the input and 
hidden layer, and among the hidden units themselves
(connecting the previous intermediate representation to the
current one), respectively, while $U$ is the output weight matrix, and
$b$ and $c$ are bias vectors connected to hidden and output units,
respectively. When $y_t$ is a scalar (hence, $U$ is a row vector) and 
$g$ is the sigmoid function, $y_t$ is simply the probability
of a positive label, conditioned on $\{x_\tau\}_{\tau=1..t}$.

For tasks requiring a single label per sequence
(e.g.\ single sentiment score per sentence), we can discard
intermediate outputs $\{y_t\}_{t=1..(T-1)}$ and use the output
of the last time step $y_T$, where $T$ is the length of the sequence.
This also means that during training, external error is only incurred
at the final time step. In general, supervision can be applied at any
intermediate time step whenever there are labels available in the dataset,
even if intermediate time step labels are not to be used at the
testing phase, since this makes
training easier.

\section{Methodology}
\subsection{Multiplicative Recurrent Neural Network}

\begin{figure*}[t!]
\centering
\includegraphics[scale=0.5]{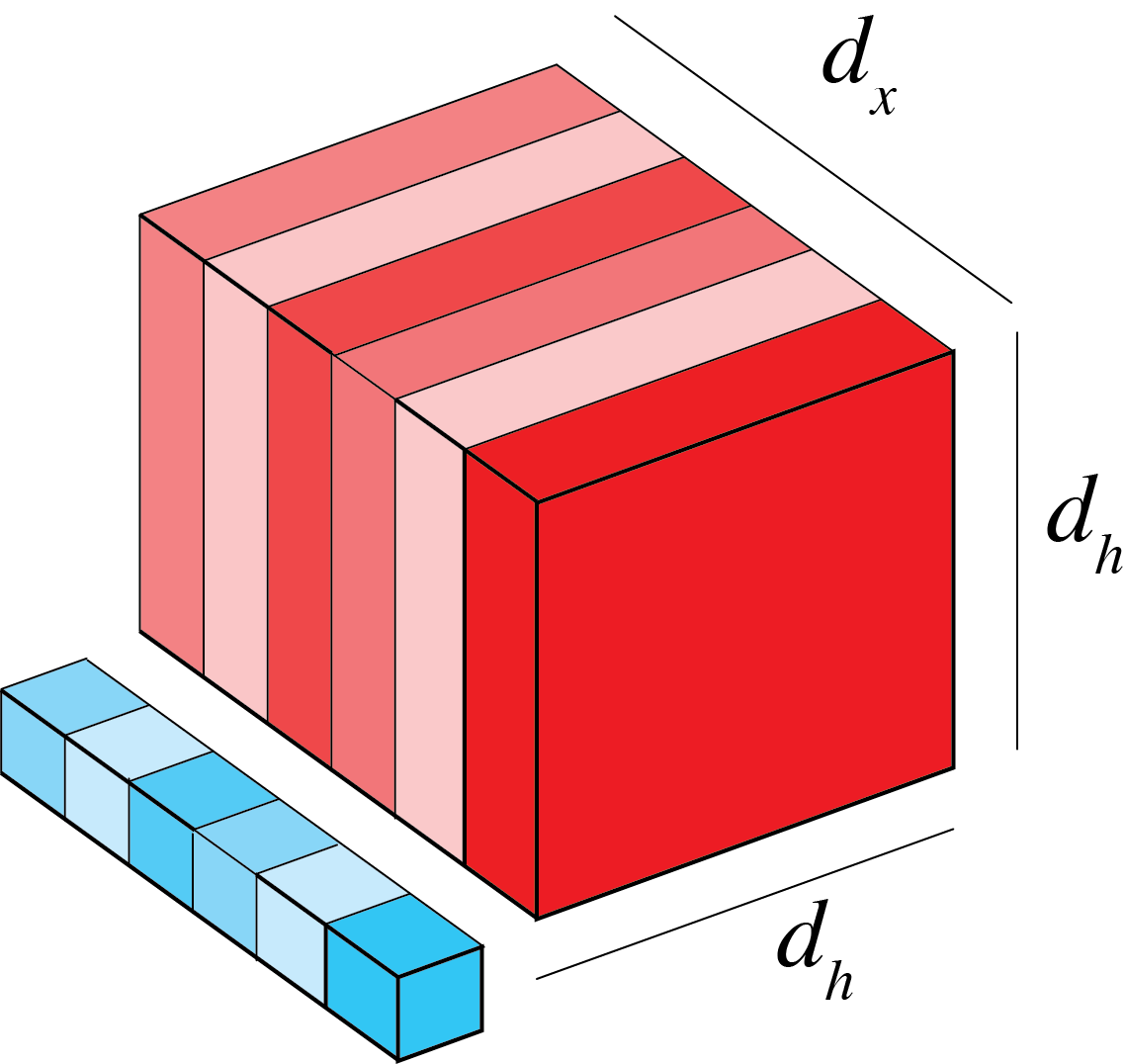}
\hspace{20pt}
\includegraphics[scale=0.5]{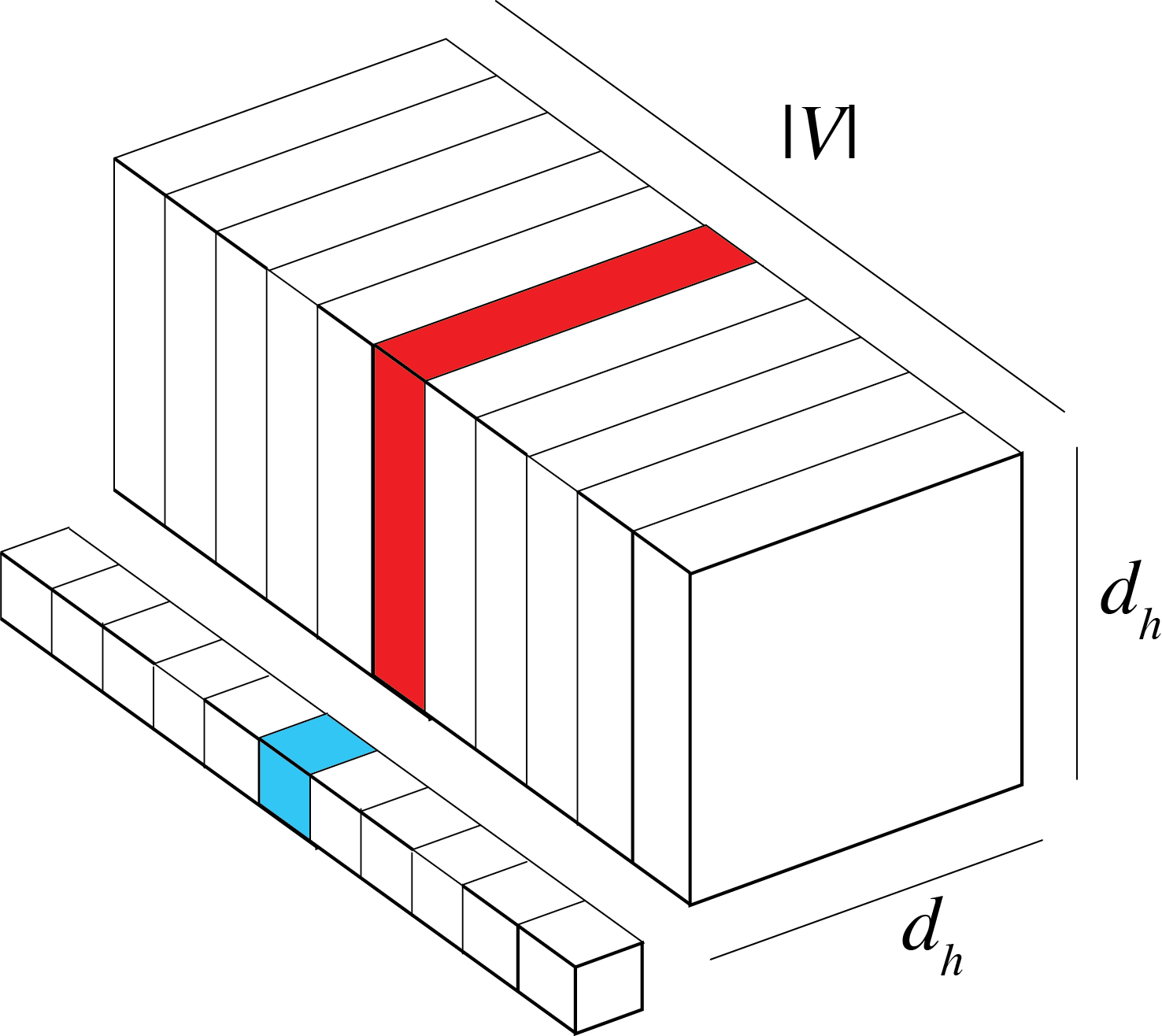}
\caption{Vector $x$ (blue) and tensor $A$ (red) sliced along the dimension
of $x$.
\textbf{Left.} Dense word vector $x$ computes a weighted sum over base
matrices to get a square matrix, which then is used to transform the meaning
vector. \textbf{Right.} One-hot word vector $x$ with the same computation,
which is equivalent to selecting one of the base matrices and falls back to
a matrix-space model. }
\label{fig:tensor}
\end{figure*}

A property of recurrent neural networks is that input layer activations
and the hidden layer activations of the previous time step
interact additively to make up the
activations for hidden layers at the current time step. This might be rather
restrictive for some applications, or difficult to learn for modeling more complex
input interactions. On the other hand, a multiplicative interaction
of those layers might provide a better representation for some semantic analysis tasks. For sentiment detection, for example, ``not'' might be considered as a 
\emph{negation} of the sentiment
that comes after it, which might be more effectively modeled with 
multiplicative interactions. To this end, we investigate 
the multiplicative recurrent neural network (or the recurrent neural tensor network)
for the sentiment analysis task that is the main focus of this paper~\citep{RNTN}.

mRNNs retain the same interpretation of memory as RNNs, the only
difference being the recursive definition of $h$:
\begin{align}
h_t &= f(x_t^\top A^{[1..d_h]} h_{t-1} + W x_t + V h_{t-1} + b) \label{eq:tensor}\\
y_t &= g(U h_t + c) \label{eq:out}
\end{align}
where $A$ is a $d_h \times d_x \times d_h$ 
tensor, and the bilinear operation $x^\top A y$ defines another vector
as $(x^\top A y)_i = x^\top A^{[i]} y$ where the right-hand side represents
the standard vector matrix multiplications and $A^{[i]}$ is a single slice
(matrix) of the tensor $A$. This means that a single entry of $h_{t,i}$
is not only a linear combination of entries $x_{t,j}$ and $h_{t-1,k}$,
but also includes multiplicative terms in the form of $a^i_{jk}x_{t,j} h_{t-1,k}$.

We can simplify Equation~\ref{eq:tensor} and \ref{eq:out} by adding bias units to $x$ and $h$:
\begin{align}
h_t &= f(x_t^{\prime\top} A'^{[1..d_h]} h'_{t-1})\\
y_t &= g(U' h'_t )
\end{align}
where $x' = [x; 1]$ and 
$h' = [h; 1]$. With this notation, $W$, $V$ and $b$
become part of the tensor $A'$ and $c$ becomes part of the matrix $U'$.

\subsection{Ordinal regression with neural networks}

\label{sec:ordinal}

Since fine-grained sentiment labels denote intensity in addition to polarity, our
class labels are ordinal in nature. Therefore, we use an ordinal regression scheme
for neural networks, as described in \citet{nnrank}. Intuitively, each sentiment class denotes
a threshold for which the instances belonging to the class have sentiment values less than
or equal to.
If an instance $s$ belongs to class $k$, it automatically belongs to the 
lower order classes
$1,\mathellipsis,k-1$, as well. Therefore, the target vector for instance $s$ is
$r = [1, \mathellipsis, 1, 0, \mathellipsis, 0]^\top$ where $r_i= 1$ if $i < k$ and $r_i=0$
otherwise. This way, we can consider the output vector as a cumulative probability distribution
on classes.

Because of the way class labels are defined, output response is not subject to normalization.
Therefore, output layer nonlinearity in this case is the elementwise sigmoid function 
$(\frac{1}{1+\exp(-x_i)})$ instead of the softmax function 
$(\frac{\exp(x_i)}{\sum_j \exp(x_j)})$ which is traditionally used for multiclass 
classification.

Note that with this scheme, output of the network is not necessarily
consistent. To decode an output vector, we firstly binarize each entry, by assigning 0
if the entry is less than 0.5 and 1 otherwise, as in conventional binary classification. 
Then we simply start from the entry with the
lowest index, and
whenever we observe a 0, we assume all of the entries with higher indices are also 0, which
ensures that the resulting target vector has the proper ordinal form. As an example,
$[1, 0, 1, 0]^\top$ is mapped to $[1, 0, 0, 0]^\top$.
Then finally, we
assign the corresponding integer label. 

\subsection{Relationship to matrix-space model}

In this section we will show the connection between mRNNs and matrix space
model.

Let us assume a purely multiplicative mRNN, without
the bias units in the input and hidden layers
(equivalently, $W = V = b = 0$).
In such an mRNN, we compute the hidden layer (memory) as follows:
\begin{align}
h_t = f(x^{\top}_t A h_{t-1})
\end{align}
Furthermore, assume $f = I$ is the identity mapping, rather than a nonlinearity function. 
We can view the tensor multiplication in two parts:
A vector $x_t$ multiplied by a tensor $A$, resulting in a matrix which we will
denote as $M(w_t)$, to make the dependence of the resulting matrix on the word $w_t$ 
explicit.
Then the matrix-vector multiplication $M(w_t) h_{t-1}$ resulting in
the vector $h_t$. Therefore, we can write the same equation as:
\begin{align}
h_t = (x^{\top}_t A) h_{t-1} = M(w_t) h_{t-1}
\end{align}
and unfolding the recursion, we have
\begin{align}
h_t = M(w_t) M(w_{t-1}) \mathellipsis M(w_1) h_0
\end{align}
If we are interested in a scalar response for the whole sequence, we apply
the output layer to the hidden layer at the final time step:
\begin{align}
y_T = u^\top h_T = u^\top M(w_T) \mathellipsis M(w_1) h_0
\end{align}
which is the matrix space model if individual $M(w_t)$ were to be
associated with the matrices of their corresponding words (Equation~\ref{eq:matspa}). 
Therefore, we can view
mRNNs as a simplification to matrix-space models in which we have a tensor
$A$ to extract a matrix for a word $w$ from its associated word vector,
rather than associating a matrix with every word. 
This can be viewed as learning
a matrix-space model with parameter sharing.
This reduces the number of
parameters greatly: instead of having a matrix for every word in the
vocabulary, we have a vector per word, and a tensor to extract matrices.

\begin{figure*}[t]
\centering
\includegraphics[scale=0.3]{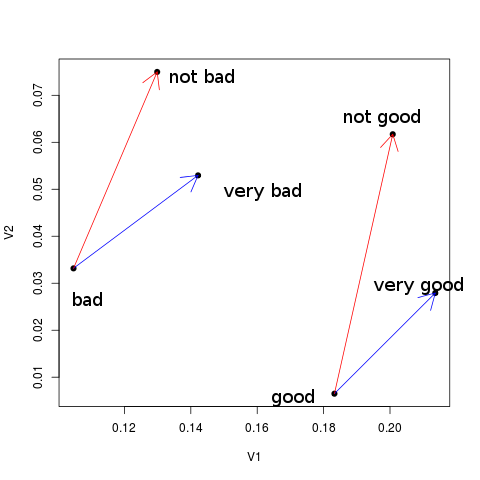}
\hspace{20pt}
\includegraphics[scale=0.3]{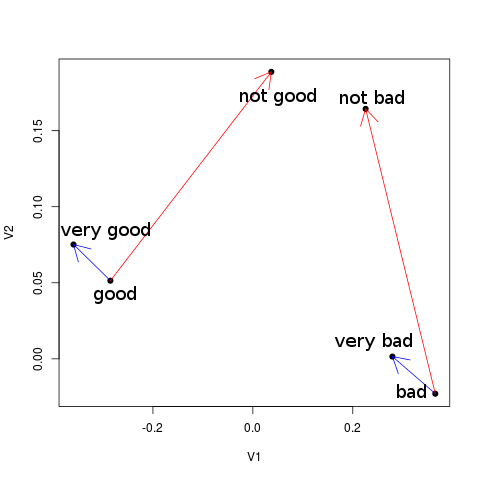}
\caption{Hidden layer vectors reduced to 2 dimensions for various phrases. \textbf{Left.}
Recurrent neural network. \textbf{Right.} Purely multiplicative recurrent neural tensor network. In mRNN,
handling of negation is more nonlinear and correctly shifts the sentiment.}
\label{fig:hidden}
\end{figure*}

Another interpretation of this is the following: instead of learning an individual
linear operator $M_w$ per word as in matrix-space models, 
mRNN learns $d_x$ number of \emph{base} linear operators. mRNN, then, 
represents each word as a weighted sum of these base operators
(weights given by the word vector $x$). Note that if $x$
is a one-hot vector representation of a word instead of a dense word embedding
(which means $d_x = |\mathcal{V}|$), then we have $|\mathcal{V}|$ matrices
as the base set of operators, and $x$ simply \emph{selects} one of these matrices,
essentially falling back to an exact matrix-space model (see Figure~\ref{fig:tensor}). 
Therefore mRNNs provide a
natural transition of the matrix-space model from a one-hot sparse word representation
to a low dimensional dense word embedding.

Besides a reduction in the number of parameters, 
another potential advantage of mRNNs over matrix-space models is that the
matrix-space model is task-dependent: for each task, one has to learn
one matrix per word in the whole vocabulary. On the other hand, mRNNs can
make use of task-independent word vectors (which can be learned in an
unsupervised manner) and only the parameters for the network would have
to be task-dependent. This allows easier extension to multitask learning
or transfer learning settings.

\section{Experiments}

\subsection{Setting}

\textbf{Data.} For experimental evaluation of the models, we use the manually 
annotated MPQA corpus \citep{wiebe2005} that contains 535 newswire documents
annotated with phrase-level subjectivity and intensity. We use the same scheme
as \citet{ainur} to preprocess and extract individual phrases from the annotated
documents, and convert the annotations to an integer ordinal label
$\{0, 1, 2, 3, 4\}$ denoting a sentiment score from negative to positive. After preprocessing,
we have 8022 phrases in total with an average length of 2.83. 
We use the training-validation-test set partitions
provided by the authors to apply 10-fold CV and report average performance over ten folds.

Additionally, we use the recently published Stanford Sentiment Treebank (SST) 
\citep{socherRNTN}, which includes labels for
215,154 phrases in the parse trees of 11,855 sentences, with an average sentence length of
19.1. Similarly, real-valued sentiment labels are converted to an integer ordinal label in
$\{0, \mathellipsis, 4\}$ by simple thresholding.
We use the single training-validation-test
set partition provided by the authors. We do not make use of the parse trees in the treebank
since our approach is not structural; however, we include the phrase-level supervised labels
(at the internal nodes of the parse trees) as labels for partial sentences.

\textbf{Problem formulation.} For experiments on the MPQA corpus, we employ
an ordinal regression setting. For experiments on SST, we employ a simple
multiclass classification setting, to make the models directly comparable to previous work.

In the classification setting, output nonlinearity $g$ is the softmax function,
and the output $y$ is a vector valued response with the class probabilities. 
Ordinal regression setting is as described in Section~\ref{sec:ordinal}.

\textbf{Evaluation metrics.} For experiments using the MPQA corpus, we use the ranking
loss as in \citet{ainur}, defined as $\frac{1}{n}\sum_i |y_i - r_i|$ where $y$ and $r$ are predicted and
true scores respectively. For experiments using SST, we use accuracy, 
$\frac{1}{n}\sum_i \mathbb{1}(y_i = r_i)$
as in \citet{socherRNTN}.

\textbf{Word vectors.} We experiment with both randomly initialized word vectors ({\sc Rand}) and
pretrained word vector representations ({\sc vec}). For pretrained word vectors, we use publicly
available 300 dimensional word vectors by \citet{mikolovnips}, trained on part of
Google News dataset ($\sim$100B words). When using pretrained word vectors, we do not finetune
them to reduce the degree of freedom of our models.

Additionally, matrix-space models are initialized with random matrices ({\sc Rand}) or a bag-of-words
regression model weights ({\sc BOW}) as described in \citet{ainur}.

\subsection{Results}

\begin{table}[t]
\parbox{.45\linewidth}{
\centering
\caption{Average ranking losses (MPQA)}
\label{tab:mpqa}
\begin{tabular}{|l|c|}
\hline
Method & Loss \\
\hline
PRank & 0.7808 \\
Bag-of-words LogReg& 0.6665 \\
Matrix-space$_{\text{Rand}}$ $(d_h = 3)$ & 0.7417 \\
Matrix-space$_{\text{BOW}}$ $(d_h = 3)$ & 0.6375 \\
\hline
RNN$^+_{\text{vec}}$ $(d_h = 315)$& 0.5265 \\
mRNN$^I_{\text{Rand}}$ $(d_h = 2)$ & 0.6799 \\
mRNN$^I_{\text{vec}}$ $(d_h = 25)$& 0.5278 \\
mRNN$^+_{\text{vec}}$ $(d_h = 25)$& 0.5232 \\
mRNN$^{\tanh}_{\text{vec}}$ $(d_h = 25)$& 0.5147 \\
\hline
\end{tabular}
} \hfill
\parbox{.45\linewidth}{
\centering
\caption{Average accuracies (SST)}
\label{tab:sst}
\begin{tabular}{|l|c|}
\hline
Method & Acc (\%) \\
\hline
Bag-of-words NB & 41.0\\
Bag-of-words SVM & 40.7\\
Bigram NB & 41.9\\
VecAvg & 32.7 \\
Recursive$^{\tanh}$ & 43.2\\
MV-Recursive$^{\tanh}$ & 44.4\\
mRecursive$^{\tanh}$ & 45.7\\
\hline
Recurrent$^+_{\text{vec}}$ $(d_h = 315)$ & 43.1\\
mRecurrent$^+_{\text{vec}}$ $(d_h = 20)$ & 43.5\\
\hline
\end{tabular}
}
\end{table}

Quantitative results on the MPQA corpus are reported in Table~\ref{tab:mpqa}. The top group shows
previous results from \citet{ainur} and the bottom group shows our results.

We observe that mRNN does slightly better that RNN with approximately the same number of parameters
(0.5232 vs.\ 0.5265). This suggests that multiplicative interactions improve the model over additive
interactions. Even though the difference is not significant in the test set, it is significant
in the development set. We partially attribute this effect to the test set variance. This also suggests that
multiplicative models are indeed more powerful, but require more careful regularization, because
early stopping with a high model variance might tend to overfit to the development set.

The randomly initialized mRNN outperforms its equivalent randomly initialized matrix-space model
(0.6799 vs.\ 0.7417), which suggests that more compact representations with shared parameters
learned by mRNN indeed
generalize better.

The mRNN and RNN that use pretrained word vectors get the best results, which suggests the importance of
good pretraining schemes, especially when supervised data is limited. This is also confirmed by our
preliminary experiments (which are not shown here)
using other word vector training methods such as CW embeddings \citep{collobert-weston} or HLBL
\citep{hlbl}, which yielded a significant difference (about $0.1-0.2$) in ranking loss.

To test the effect of different nonlinearities, we experiment with the identity, rectifier and 
$\tanh$ functions with mRNNs. Experiments show that there is small but consistent improvement
as we use rectifier or $\tanh$ over not using extra nonlinearity. The differences between
rectifier and identity, and $\tanh$ and rectifier are not significant; however, the difference
between $\tanh$ and identity is significant, suggesting a performance boost from using a nonlinear
squashing function.
Nonetheless, not using any nonlinearity is only marginally worse. A possible explanation is that since
the squashing function is not the only source of nonlinearity in mRNNs (multiplicativeness
is another source of nonlinearity), it is not as crucial.

Results on the Stanford Sentiment Treebank are shown in Table~\ref{tab:sst}. Again, the top group shows
baselines from \citet{socherRNTN} and the bottom group shows our results.

Both RNN and mRNN outperform the conventional SVM and Naive Bayes baselines. We observe that RNN
can get very close to the performance of Recursive Neural Network, which can be considered
its structural counterpart. mRNN further improves over RNN and performs better than the recursive net
and worse than the matrix-vector recursive net. Note that none of the RNN-based methods employ parse trees
of sentences, unlike their recursive neural network variants.

\section{Conclusion and Discussion}

In this work, we explore multiplicative recurrent neural networks as a model for the 
compositional interpretation of language. We evaluate on the task of fine-grained sentiment analysis, in an ordinal regression setting and show that mRNNs outperform previous work
on MPQA, and get comparable results to previous work on Stanford Sentiment Treebank without using
parse trees.  We also describe how mRNNs effectively generalize matrix-space models from a sparse 1-hot word vector
representation to a distributed, dense representation. 

One benefit of mRNNs over matrix-space models
is their separation of task-independent word representations (vectors) from task-dependent classifiers
(tensor), making them very easy to extend for semi-supervised learning or transfer learning settings.
Slices of the tensor can be interpreted as \emph{base matrices} of a simplified matrix-space model. 
Intuitively, every \emph{meaning factor} (a dimension of the dense word vector) of a word has a separate operator
acting on the meaning representation which we combine to get the operator of the word itself. 

From a parameter
sharing perspective, mRNNs provide better models. For matrix-space models, an update over a sentence affects only the word
matrices that occur in that particular sentence. On the other hand, in an mRNN, an update over a sentence affects
the global tensor as well. With such an update, the network alters its operation for similar words towards a similar direction.

One drawback of mRNNs over conventional additive RNNs is their increased model variance, resulting from multiplicative interactions.
This can be tackled by a stricter regularization. Another future direction is to explore sparsity constraints on word vectors, which would mean that
every word would select only a few base operators to act on the meaning representation.

\subsubsection*{Acknowledgments}
This work was supported in part by NSF grant IIS-1314778 and DARPA DEFT Grant FA8750-13-2-0015.
The views and conclusions contained herein are those of the authors
and should not be interpreted as necessarily representing the official
policies or endorsements, either expressed or implied, of NSF, DARPA
or the U.S.\ Government.

{\small
\bibliography{ref}

\begin{thebibliography}{29}
\providecommand{\natexlab}[1]{#1}
\providecommand{\url}[1]{\texttt{#1}}
\expandafter\ifx\csname urlstyle\endcsname\relax
  \providecommand{\doi}[1]{doi: #1}\else
  \providecommand{\doi}{doi: \begingroup \urlstyle{rm}\Url}\fi

\bibitem[Baroni \& Zamparelli(2010)Baroni and Zamparelli]{baroni2010}
Baroni, Marco and Zamparelli, Roberto.
\newblock Nouns are vectors, adjectives are matrices: Representing
  adjective-noun constructions in semantic space.
\newblock In \emph{Proceedings of the 2010 Conference on Empirical Methods in
  Natural Language Processing}, pp.\  1183--1193. Association for Computational
  Linguistics, 2010.

\bibitem[Bengio et~al.(2001)Bengio, Ducharme, Vincent, Jauvin, K, Hofmann,
  Poggio, and Shawe-taylor]{bengioNeural}
Bengio, Yoshua, Ducharme, Réjean, Vincent, Pascal, Jauvin, Christian, K, Jaz,
  Hofmann, Thomas, Poggio, Tomaso, and Shawe-taylor, John.
\newblock A neural probabilistic language model.
\newblock In \emph{In Advances in Neural Information Processing Systems}, 2001.

\bibitem[Cheng et~al.(2008)Cheng, Wang, and Pollastri]{nnrank}
Cheng, Jianlin, Wang, Zheng, and Pollastri, Gianluca.
\newblock A neural network approach to ordinal regression.
\newblock In \emph{Neural Networks, 2008. IJCNN 2008.(IEEE World Congress on
  Computational Intelligence). IEEE International Joint Conference on}, pp.\
  1279--1284. IEEE, 2008.

\bibitem[Clark(2008)]{clark}
Clark, Stephen.
\newblock A compositional distributional model of meaning.
\newblock In \emph{Proceedings of the Second Quantum Interaction Symposium
  (QI-2008)}, 2008.

\bibitem[Coecke et~al.(2010)Coecke, Sadrzadeh, and S.]{coecke}
Coecke, B., Sadrzadeh, M., and S., Clark.
\newblock Mathematical foundations for a compositional distributional model of
  meaning.
\newblock \emph{Linguistic Analysis}, 36:\penalty0 345--384, 2010.

\bibitem[Collobert \& Weston(2008)Collobert and Weston]{collobert-weston}
Collobert, Ronan and Weston, Jason.
\newblock A unified architecture for natural language processing: Deep neural
  networks with multitask learning.
\newblock In \emph{Proceedings of the 25th international conference on Machine
  learning}, pp.\  160--167. ACM, 2008.

\bibitem[Collobert et~al.(2011)Collobert, Weston, Bottou, Karlen, Kavukcuoglu,
  and Kuksa]{collobertScratch}
Collobert, Ronan, Weston, Jason, Bottou, L{\'e}on, Karlen, Michael,
  Kavukcuoglu, Koray, and Kuksa, Pavel.
\newblock Natural language processing (almost) from scratch.
\newblock \emph{J. Mach. Learn. Res.}, 12:\penalty0 2493--2537, November 2011.
\newblock ISSN 1532-4435.
\newblock URL \url{http://dl.acm.org/citation.cfm?id=1953048.2078186}.

\bibitem[Elman(1990)]{elman}
Elman, Jeffrey~L.
\newblock Finding structure in time.
\newblock \emph{Cognitive science}, 14\penalty0 (2):\penalty0 179--211, 1990.

\bibitem[Grefenstette et~al.(2013)Grefenstette, Dinu, Zhang, Sadrzadeh, and
  Baroni]{grefenstette-et-al:2013:IWCS2013}
Grefenstette, E., Dinu, G., Zhang, Y., Sadrzadeh, M., and Baroni, M.
\newblock Multi-step regression learning for compositional distributional
  semantics.
\newblock In \emph{Proceedings of the 10th International Conference on
  Computational Semantics (IWCS 2013) -- Long Papers}, pp.\  131--142, Potsdam,
  Germany, March 2013. Association for Computational Linguistics.
\newblock URL \url{http://www.aclweb.org/anthology/W13-0112}.

\bibitem[Kennedy \& Inkpen(2006)Kennedy and Inkpen]{kennedy}
Kennedy, Alistair and Inkpen, Diana.
\newblock Sentiment classification of movie reviews using contextual valence
  shifters.
\newblock \emph{Computational Intelligence}, 22\penalty0 (2):\penalty0
  110--125, 2006.

\bibitem[Mikolov et~al.(2013{\natexlab{a}})Mikolov, Chen, Corrado, and
  Dean]{mikolov}
Mikolov, Tomas, Chen, Kai, Corrado, Greg, and Dean, Jeffrey.
\newblock Efficient estimation of word representations in vector space.
\newblock \emph{arXiv preprint arXiv:1301.3781}, 2013{\natexlab{a}}.

\bibitem[Mikolov et~al.(2013{\natexlab{b}})Mikolov, Sutskever, Chen, Corrado,
  and Dean]{mikolovnips}
Mikolov, Tomas, Sutskever, Ilya, Chen, Kai, Corrado, Greg~S, and Dean, Jeff.
\newblock Distributed representations of words and phrases and their
  compositionality.
\newblock In \emph{Advances in Neural Information Processing Systems}, pp.\
  3111--3119, 2013{\natexlab{b}}.

\bibitem[Mitchell \& Lapata(2010)Mitchell and Lapata]{Mitchell:Lapata:2010}
Mitchell, Jeff and Lapata, Mirella.
\newblock Composition in distributional models of semantics.
\newblock \emph{Cognitive Science}, 34\penalty0 (8):\penalty0 1388--1439, 2010.

\bibitem[Mnih \& Hinton(2007)Mnih and Hinton]{hlbl}
Mnih, Andriy and Hinton, Geoffrey.
\newblock Three new graphical models for statistical language modelling.
\newblock In \emph{Proceedings of the 24th international conference on Machine
  learning}, pp.\  641--648. ACM, 2007.

\bibitem[Pang \& Lee(2008)Pang and Lee]{pang2008}
Pang, Bo and Lee, Lillian.
\newblock Opinion mining and sentiment analysis.
\newblock \emph{Foundations and trends in information retrieval}, 2\penalty0
  (1-2):\penalty0 1--135, 2008.

\bibitem[Polanyi \& Zaenen(2006)Polanyi and Zaenen]{polanyi}
Polanyi, Livia and Zaenen, Annie.
\newblock Contextual valence shifters.
\newblock In \emph{Computing attitude and affect in text: Theory and
  applications}, pp.\  1--10. Springer, 2006.

\bibitem[Pollack(1990)]{pollack:raam}
Pollack, J.~B.
\newblock Recursive distributed representations.
\newblock \emph{Artificial Intelligence}, 1:\penalty0 77--105, 1990.

\bibitem[Rudolph \& Giesbrecht(2010)Rudolph and Giesbrecht]{rudolph2010}
Rudolph, Sebastian and Giesbrecht, Eugenie.
\newblock Compositional matrix-space models of language.
\newblock In \emph{Proceedings of the 48th Annual Meeting of the Association
  for Computational Linguistics}, pp.\  907--916. Association for Computational
  Linguistics, 2010.

\bibitem[Shaikh et~al.(2007)Shaikh, Prendinger, and Mitsuru]{shaikh}
Shaikh, Mostafa Al~Masum, Prendinger, Helmut, and Mitsuru, Ishizuka.
\newblock Assessing sentiment of text by semantic dependency and contextual
  valence analysis.
\newblock In \emph{Affective Computing and Intelligent Interaction}, pp.\
  191--202. Springer, 2007.

\bibitem[Socher et~al.(2011)Socher, Lin, Ng, and Manning]{socher2011parsing}
Socher, Richard, Lin, Cliff~C, Ng, Andrew, and Manning, Chris.
\newblock Parsing natural scenes and natural language with recursive neural
  networks.
\newblock In \emph{Proceedings of the 28th International Conference on Machine
  Learning (ICML-11)}, pp.\  129--136, 2011.

\bibitem[Socher et~al.(2012)Socher, Huval, Manning, and Ng]{socherMVRNN}
Socher, Richard, Huval, Brody, Manning, Christopher~D, and Ng, Andrew~Y.
\newblock Semantic compositionality through recursive matrix-vector spaces.
\newblock In \emph{Proceedings of the 2012 Joint Conference on Empirical
  Methods in Natural Language Processing and Computational Natural Language
  Learning}, pp.\  1201--1211. Association for Computational Linguistics, 2012.

\bibitem[Socher et~al.(2013)Socher, Perelygin, Wu, Chuang, Manning, Ng, and
  Potts]{socherRNTN}
Socher, Richard, Perelygin, Alex, Wu, Jean~Y, Chuang, Jason, Manning,
  Christopher~D, Ng, Andrew~Y, and Potts, Christopher.
\newblock Recursive deep models for semantic compositionality over a sentiment
  treebank.
\newblock In \emph{Proceedings of the Conference on Empirical Methods in
  Natural Language Processing}, EMNLP '13, 2013.

\bibitem[Sutskever et~al.(2011)Sutskever, Martens, and Hinton]{RNTN}
Sutskever, Ilya, Martens, James, and Hinton, Geoffrey~E.
\newblock Generating text with recurrent neural networks.
\newblock In \emph{Proceedings of the 28th International Conference on Machine
  Learning (ICML-11)}, pp.\  1017--1024, 2011.

\bibitem[Turian et~al.(2010)Turian, Ratinov, and Bengio]{turian}
Turian, Joseph, Ratinov, Lev, and Bengio, Yoshua.
\newblock Word representations: a simple and general method for semi-supervised
  learning.
\newblock In \emph{Proceedings of the 48th Annual Meeting of the Association
  for Computational Linguistics}, pp.\  384--394. Association for Computational
  Linguistics, 2010.

\bibitem[Widdows(2003)]{widdows:2003:ACL}
Widdows, Dominic.
\newblock Orthogonal negation in vector spaces for modelling word-meanings and
  document retrieval.
\newblock In \emph{Proceedings of the 41st Annual Meeting of the Association
  for Computational Linguistics}, pp.\  136--143, Sapporo, Japan, July 2003.
  Association for Computational Linguistics.
\newblock \doi{10.3115/1075096.1075114}.
\newblock URL \url{http://www.aclweb.org/anthology/P03-1018}.

\bibitem[Wiebe et~al.(2005)Wiebe, Wilson, and Cardie]{wiebe2005}
Wiebe, Janyce, Wilson, Theresa, and Cardie, Claire.
\newblock Annotating expressions of opinions and emotions in language.
\newblock \emph{Language resources and evaluation}, 39\penalty0 (2-3):\penalty0
  165--210, 2005.

\bibitem[Wilson et~al.(2005)Wilson, Wiebe, and Hoffmann]{wilson2005}
Wilson, Theresa, Wiebe, Janyce, and Hoffmann, Paul.
\newblock Recognizing contextual polarity in phrase-level sentiment analysis.
\newblock In \emph{Proceedings of the Conference on Human Language Technology
  and Empirical Methods in Natural Language Processing}, pp.\  347--354.
  Association for Computational Linguistics, 2005.

\bibitem[Yessenalina \& Cardie(2011)Yessenalina and Cardie]{ainur}
Yessenalina, Ainur and Cardie, Claire.
\newblock Compositional matrix-space models for sentiment analysis.
\newblock In \emph{Proceedings of the Conference on Empirical Methods in
  Natural Language Processing}, pp.\  172--182. Association for Computational
  Linguistics, 2011.

\bibitem[Zanzotto et~al.(2010)Zanzotto, Korkontzelos, Fallucchi, and
  Manandhar]{zanzotto-EtAl:2010:PAPERS}
Zanzotto, Fabio~Massimo, Korkontzelos, Ioannis, Fallucchi, Francesca, and
  Manandhar, Suresh.
\newblock Estimating linear models for compositional distributional semantics.
\newblock In \emph{Proceedings of the 23rd International Conference on
  Computational Linguistics (Coling 2010)}, pp.\  1263--1271, Beijing, China,
  August 2010. Coling 2010 Organizing Committee.
\newblock URL \url{http://www.aclweb.org/anthology/C10-1142}.

\end{thebibliography}
}
\bibliographystyle{iclr2015}

\end{document}